\documentclass[sigconf]{acmart}
\AtBeginDocument{%
  }

\setcopyright{acmlicensed}
\copyrightyear{2025}
\acmYear{2025}
\acmDOI{}
\acmConference{Causality, Counterfactuals \& Sequential Decision-Making Workshop at RecSys '25}{September 22,
  2025}{Prague, Czech Republic}

\begin{document}

\title{Counterfactual Risk Minimization with IPS-Weighted BPR and Self-Normalized Evaluation in Recommender Systems}

\author{Rahul Raja}
\authornote{This work does not relate to  
 Position at LinkedIn.}
\affiliation{%
  \institution{Carnegie Mellon University, Linkedin}
  \country{USA}
}

\author{Arpita Vats}
\authornotemark[1]
\affiliation{%
  \institution{Boston University, Linkedin}
  \country{USA}}









\begin{abstract}
Learning and evaluating recommender systems from logged implicit feedback is challenging due to exposure bias. While inverse propensity scoring (IPS) corrects this bias, it often suffers from high variance and instability. In this paper, we present a simple and effective pipeline that integrates IPS-weighted training with an IPS-weighted Bayesian Personalized Ranking (BPR) objective augmented by a Propensity Regularizer (PR). We compare Direct Method (DM), IPS, and Self-Normalized IPS (SNIPS) for offline policy evaluation, and demonstrate how IPS-weighted training improves model robustness under biased exposure. The proposed PR further mitigates variance amplification from extreme propensity weights, leading to more stable estimates. Experiments on synthetic and MovieLens 100K data show that our approach generalizes better under unbiased exposure while reducing evaluation variance compared to naive and standard IPS methods, offering practical guidance for counterfactual learning and evaluation in real-world recommendation settings
\end{abstract}




\received{20 February 2007}
\received[revised]{12 March 2009}
\received[accepted]{5 June 2009}

\maketitle

\section{Introduction}

Recommender systems are essential for enabling users to navigate vast content catalogs in domains such as e-commerce, video streaming, and social media~\cite{ricci2011introduction,zhang2019deep}. These systems are often trained and evaluated using \emph{implicit feedback} logs, such as clicks, views, or watch time, where feedback is only observed for items that were exposed to the user. This leads to \textbf{exposure bias}~\cite{joachims2017unbiased}, since unexposed items have no opportunity to receive feedback, making both learning and evaluation biased toward the logging policy.  

A principled approach to correct for this bias is \textbf{Inverse Propensity Scoring (IPS)}~\cite{schnabel2016recommendations,swaminathan2015batch}, which reweights observed data by the inverse of the logging propensity. IPS has been widely applied in counterfactual learning and off-policy evaluation~\cite{bottou2013counterfactual}, but its practical use is hindered by high variance when propensities are small. To mitigate this, \textbf{Self-Normalized IPS (SNIPS)}~\cite{swaminathan2015counterfactual} normalizes the weights, reducing variance at the expense of introducing a small bias.  

Recent work has primarily explored IPS and SNIPS in isolation, often in the context of matrix factorization~\cite{schnabel2016recommendations, yao2021debiased}. However, their combined application as a lightweight, unified pipeline for both \emph{training} and \emph{evaluation} in modern graph-based recommenders remains underexplored. In particular, Light Graph Convolutional Networks (LightGCN)~\cite{he2020lightgcn}, which achieve state-of-the-art performance in collaborative filtering, have not been systematically studied under counterfactual learning with IPS-weighted objectives.

In this paper, we address this gap by proposing a simple yet effective training–evaluation pipeline that integrates IPS into the BPR loss~\cite{rendle2009bpr} and augments it with a \textbf{Propensity Regularizer (PR)} to control the variance amplification caused by extreme propensity weights. For evaluation, we compare Direct Method (DM), IPS, and SNIPS estimators, providing a systematic empirical study of their trade-offs.Our main contributions are:
\begin{itemize}
    \item We implement \textbf{IPS-weighted BPR loss} in a LightGCN model, enabling debiased learning from logged implicit feedback while preserving pairwise ranking optimization.
    \item We introduce a \textbf{Propensity Regularizer} that penalizes large IPS weights during training, mitigating the high variance issue inherent to inverse propensity weighting.
    \item We perform a comprehensive empirical comparison of \emph{Direct Method}, \emph{IPS}, and \emph{SNIPS} estimators for offline policy evaluation on both synthetic and real-world (MovieLens 100K) datasets.
    \item We show that our approach improves generalization under unbiased exposure and produces more stable evaluation estimates than naive and standard IPS baselines.
\end{itemize}

Our results highlight that combining IPS-weighted training with SNIPS evaluation is a practical and effective counterfactual learning strategy for graph-based recommender systems in real-world settings.

\section{Background and Related Work}
\label{sec:background}

\subsection{Exposure Bias in Implicit Feedback}
In recommender systems, user feedback is often \emph{implicit} (e.g., clicks, views, watch time), which only records interactions for items that were exposed to the user. This leads to \textbf{exposure bias}~\cite{joachims2017unbiased, schnabel2016recommendations}, since the absence of interaction does not necessarily imply negative preference. Let $\mathcal{D} = \{ (u, i, r_{ui}, b_{ui}) \}$ denote the logged dataset, where $r_{ui} \in \{0,1\}$ is the observed feedback and $b_{ui}$ is the propensity, i.e., the probability that item $i$ was shown to user $u$. A naive empirical risk minimization objective over $\mathcal{D}$ implicitly assumes $b_{ui}=1$ for all $(u,i)$, which biases both learning and evaluation.

\subsection{Inverse Propensity Scoring (IPS)}
Inverse Propensity Scoring~\cite{bottou2013counterfactual, swaminathan2015counterfactual} provides an unbiased estimate of the target policy’s expected reward by reweighting logged interactions according to their inverse propensities:
\begin{equation}
    \hat{R}_{\text{IPS}} = \frac{1}{|\mathcal{D}|} \sum_{(u,i) \in \mathcal{D}} \frac{\pi(i|u)}{b_{ui}} \, r_{ui}
    \label{eq:ips}
\end{equation}
where $\pi(i|u)$ is the target policy's probability of recommending $i$ to $u$. While IPS is unbiased, it can suffer from high variance when $b_{ui}$ is small, as large weights $\frac{1}{b_{ui}}$ can dominate the estimate.

In learning, the IPS principle can be integrated into a loss function. For example, for binary cross-entropy (BCE) loss:
\begin{equation}
    \mathcal{L}_{\text{IPS-BCE}} = -\sum_{(u,i) \in \mathcal{D}} \frac{1}{b_{ui}} \left[ r_{ui} \log \hat{y}_{ui} + (1 - r_{ui}) \log (1 - \hat{y}_{ui}) \right]
    \label{eq:ips_bce}
\end{equation}
where $\hat{y}_{ui}$ is the model’s predicted probability.

\subsection{Self-Normalized IPS (SNIPS)}
Self-Normalized IPS~\cite{swaminathan2015batch} addresses the variance issue by normalizing the weights:
\begin{equation}
    \hat{R}_{\text{SNIPS}} = \frac{\sum_{(u,i) \in \mathcal{D}} \frac{\pi(i|u)}{b_{ui}} r_{ui}}{\sum_{(u,i) \in \mathcal{D}} \frac{\pi(i|u)}{b_{ui}}}
    \label{eq:snips}
\end{equation}
This reduces variance at the cost of introducing a small bias, but is particularly effective in offline evaluation where stability is important.

\subsection{LightGCN for Recommendation}
LightGCN~\cite{he2020lightgcn} is a simplified Graph Convolutional Network for collaborative filtering, designed to eliminate feature transformation and non-linearities. Given a bipartite user–item graph $\mathcal{G} = (\mathcal{U} \cup \mathcal{I}, \mathcal{E})$ with normalized adjacency matrix $\tilde{A}$, LightGCN computes embeddings by layer-wise propagation:
\begin{equation}
    \mathbf{e}_u^{(k+1)} = \sum_{i \in \mathcal{N}_u} \frac{1}{\sqrt{|\mathcal{N}_u|}\sqrt{|\mathcal{N}_i|}} \, \mathbf{e}_i^{(k)}, \quad
    \mathbf{e}_i^{(k+1)} = \sum_{u \in \mathcal{N}_i} \frac{1}{\sqrt{|\mathcal{N}_i|}\sqrt{|\mathcal{N}_u|}} \, \mathbf{e}_u^{(k)}
\end{equation}
Final embeddings are an average over all propagation layers:
\begin{equation}
    \mathbf{e}_u = \frac{1}{K+1} \sum_{k=0}^K \mathbf{e}_u^{(k)}, \quad \mathbf{e}_i = \frac{1}{K+1} \sum_{k=0}^K \mathbf{e}_i^{(k)}
\end{equation}
The predicted score is $\hat{y}_{ui} = \mathbf{e}_u^\top \mathbf{e}_i$.

\subsection{Bias-Aware Learning Objectives}
While LightGCN has shown strong performance in unbiased datasets, its training objective can be adapted to handle exposure bias. Bias-aware approaches include:
\begin{itemize}
    \item \textbf{IPS-weighted BCE or BPR}: Modifying the loss with $1/b_{ui}$ weights~\cite{yao2021debiased}.
    \item \textbf{Propensity Regularization (PR)}: Adding a regularizer to prevent overfitting to rare exposures.
    \item \textbf{Self-normalization in evaluation}: Using SNIPS in Eq.~\ref{eq:snips} to stabilize offline metrics.
\end{itemize}
Our work differs from prior methods~\cite{joachims2017unbiased, yao2021debiased} by combining IPS-weighted LightGCN training with SNIPS evaluation in a unified pipeline, empirically analyzing their joint effectiveness on biased implicit feedback datasets.

\section{Methodology and Experimental Setup}

\subsection{Problem Formulation}
We consider a recommendation scenario with a set of users $\mathcal{U}$ and items $\mathcal{I}$. The user–item interaction matrix $R \in \{0,1\}^{|\mathcal{U}| \times |\mathcal{I}|}$ contains implicit feedback, where $r_{ui}=1$ indicates that user $u$ interacted with item $i$. Observations are limited to exposed items, represented by an exposure indicator $o_{ui} \in \{0,1\}$, leading to \textit{exposure bias}.

Let $b(u,i)$ denote the logging policy's propensity of exposing $(u,i)$, and $\pi(u,i)$ denote the target policy's probability of recommending $(u,i)$. The IPS-weighted loss for binary cross-entropy (BCE) training is:
\[
\mathcal{L}_{\text{IPS}} = - \sum_{(u, i) \in \mathcal{O}} \frac{\pi(u, i)}{b(u, i)} \cdot \big[ r_{ui} \log \sigma(\hat{y}_{ui}) + (1-r_{ui}) \log (1-\sigma(\hat{y}_{ui})) \big],
\]
where $\sigma$ is the sigmoid function and $\mathcal{O}$ is the set of observed interactions.  
For Bayesian Personalized Ranking (BPR), we apply:
\[
\mathcal{L}_{\text{IPS-BPR}} = - \sum_{(u, i, j) \in \mathcal{D}} \frac{\pi(u, i)}{b(u, i)} \cdot \log \sigma(\hat{y}_{ui} - \hat{y}_{uj}),
\]
optionally including a propensity regularization (PR) term to prevent extreme weight amplification.

\subsection{Evaluation with SNIPS}
To reduce the high variance of IPS, we employ the Self-Normalized IPS estimator:
\[
\hat{V}_{\text{SNIPS}} = \frac{\sum_{(u, i) \in \mathcal{O}} \frac{\pi(u, i)}{b(u, i)} r_{ui}}{\sum_{(u, i) \in \mathcal{O}} \frac{\pi(u, i)}{b(u, i)}}.
\]
We also compute the Effective Sample Size (ESS) to quantify the variance–bias tradeoff.

\subsection{Model Architecture}
We use \textbf{LightGCN}~\cite{he2020lightgcn} as the base recommender. The model propagates user/item embeddings through $K$ layers via:
\[
\mathbf{e}^{(k+1)} = \tilde{\mathbf{A}} \mathbf{e}^{(k)},
\]
where $\tilde{\mathbf{A}}$ is the symmetrically normalized adjacency matrix of the user–item graph. The final embedding is the layer-wise average:
\[
\mathbf{e}_u = \frac{1}{K+1} \sum_{k=0}^K \mathbf{e}_u^{(k)}.
\]

\subsection{Experimental Setup}
We evaluate on:
\begin{itemize}
    \item \textbf{MovieLens 100K} dataset, where exposure bias is simulated by sampling exposures with a softmax over item popularity, controlled by a bias parameter.
    \item \textbf{Synthetic dataset} generated to verify behavior under controlled bias conditions(explained in Appendix).
\end{itemize}
We compare Naive, IPS, and IPS-BPR+PR training strategies, all evaluated with SNIPS.  
Hyperparameters: embedding size $64$, layers $K=3$, Adam optimizer with learning rate $10^{-3}$, batch size $1024$, and early stopping on validation NDCG.  
Variance estimates are computed via 50 bootstrap resamples per run.

\section{Results and Discussion}

We evaluate the performance of our proposed \textit{IPS-weighted BPR + Propensity Regularizer (PR)} approach against baseline methods, focusing on its ability to mitigate exposure bias and improve offline policy evaluation accuracy.

\subsection{Estimated Reward Distributions}
\begin{figure}[htbp]
    \centering
    \includegraphics[width=0.48\textwidth]{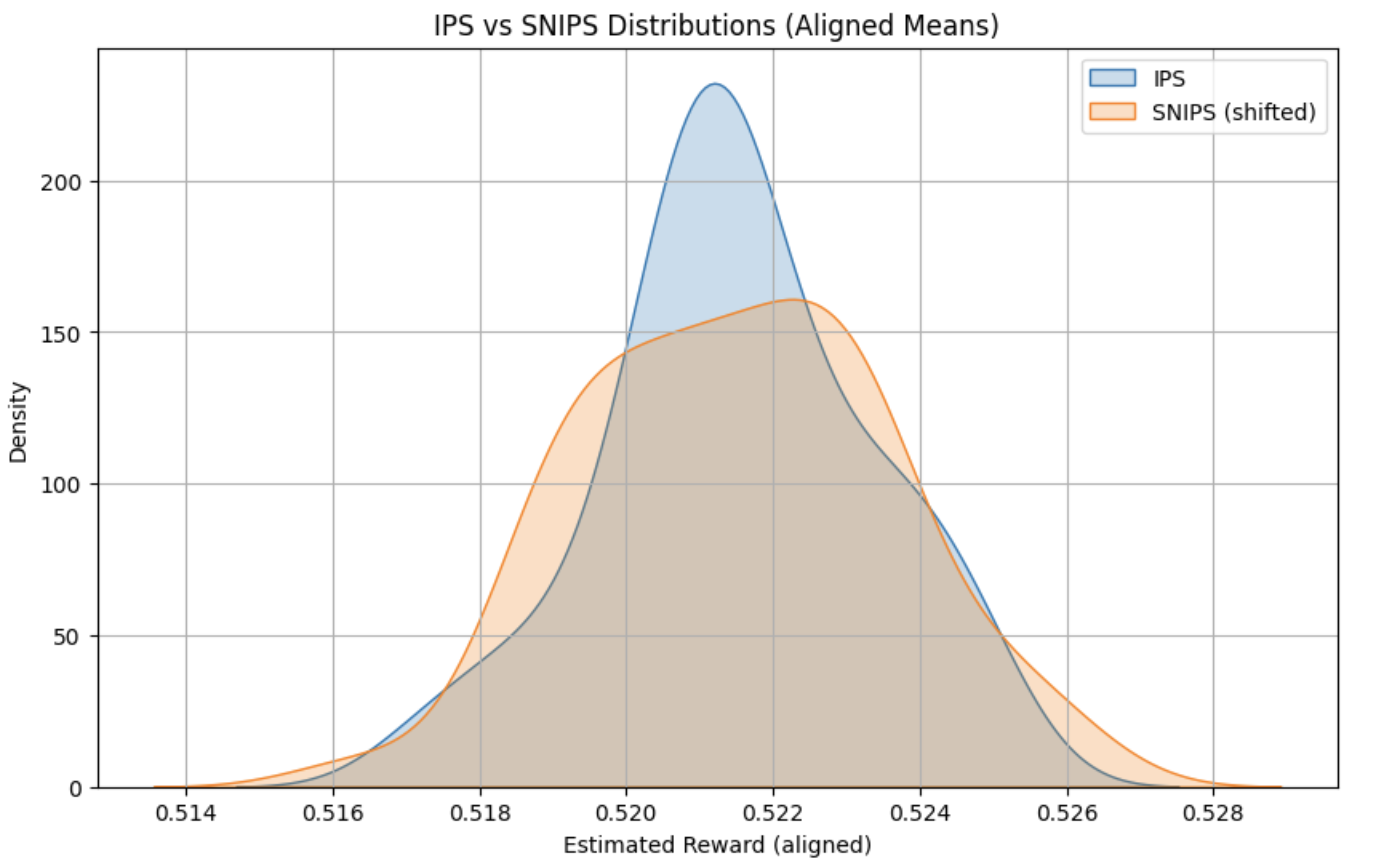}
    \caption{IPS vs SNIPS estimated reward distributions after mean alignment. SNIPS exhibits reduced variance compared to IPS.}
    \label{fig:reward_dist}
\end{figure}

Figure~\ref{fig:reward_dist} compares the estimated reward distributions for IPS and SNIPS estimators after aligning their means. We observe that SNIPS produces a smoother, slightly shifted distribution compared to IPS, indicating its normalization effect, which reduces variance at the cost of introducing mild bias.

\subsection{Sensitivity to Exposure Bias Level}
\begin{figure}[htbp]
    \centering
    \includegraphics[width=0.48\textwidth]{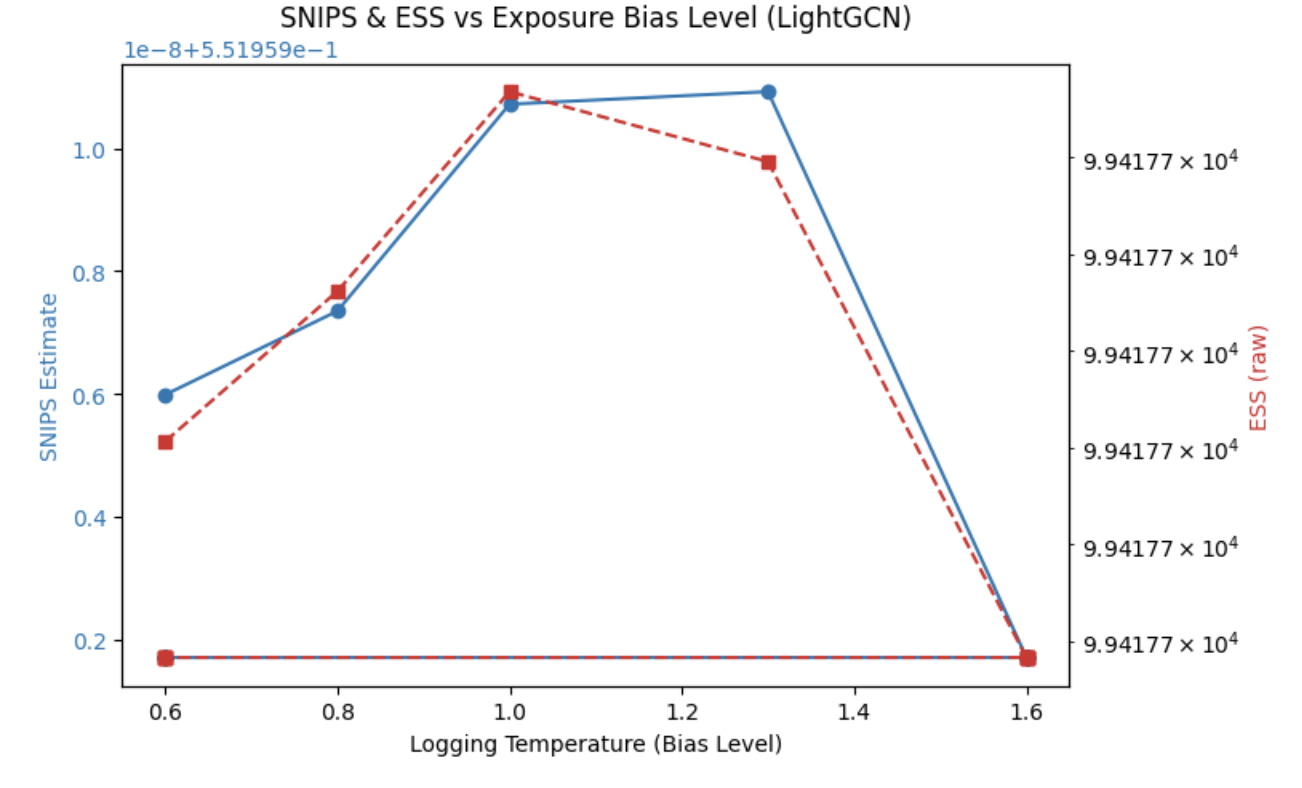}
    \caption{SNIPS and ESS across different exposure bias levels. Optimal performance is observed at moderate bias.}
    \label{fig:bias_level}
\end{figure}

Figure~\ref{fig:bias_level} illustrates the relationship between SNIPS estimates, Effective Sample Size (ESS), and the logging temperature parameter, which controls exposure bias. We find that moderate bias levels (temperature $\approx 1.0$--$1.2$) yield the highest SNIPS estimates and stable ESS, while extreme bias settings (low or high temperature) result in degraded performance due to poor coverage.

\subsection{Learning Curves Across BPR Variants}
\begin{figure}[htbp]
    \centering
    \includegraphics[width=0.48\textwidth]{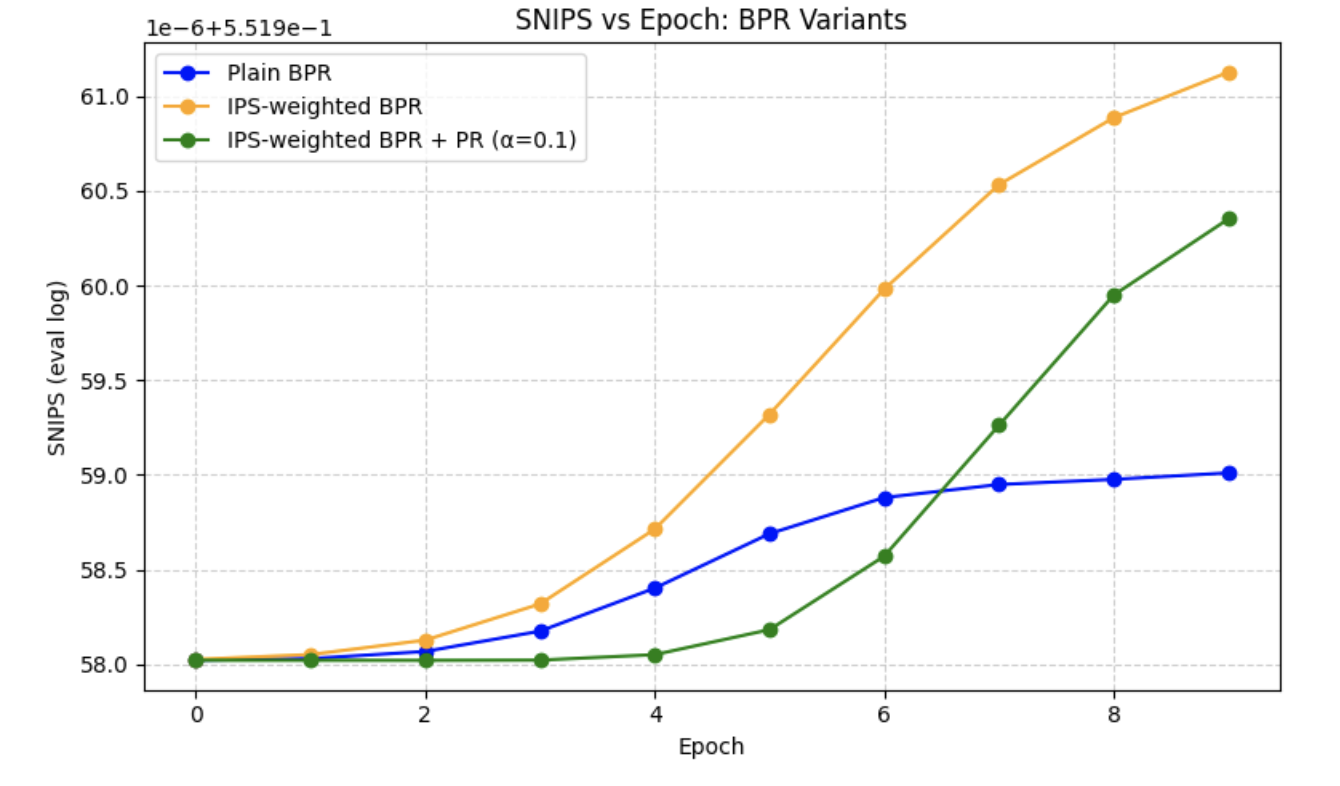}
    \caption{SNIPS evaluation across epochs for BPR variants. IPS-weighted BPR + PR achieves stable improvements over the baseline.}
    \label{fig:eval_log}
\end{figure}

Figure~\ref{fig:eval_log} presents the SNIPS-evaluated learning curves for three BPR variants: Plain BPR, IPS-weighted BPR, and IPS-weighted BPR with PR ($\alpha=0.1$). The IPS-weighted variants consistently outperform the plain BPR baseline, demonstrating the effectiveness of importance weighting. The addition of the propensity regularizer leads to more stable early-stage training and competitive final performance, especially when considering generalization under exposure bias.

\subsection{Discussion}
Our results show that:
\begin{itemize}
    \item \textbf{Variance Reduction:} SNIPS consistently lowers estimator variance compared to IPS (Figure~\ref{fig:reward_dist}), which is critical in high-bias settings where IPS becomes unstable.
    \item \textbf{Bias Sensitivity:} Performance peaks at moderate exposure bias (Figure~\ref{fig:bias_level}); too little bias limits gains from debiasing, while too much bias reduces effective sample size.
    \item \textbf{Regularization Benefits:} The propensity regularizer improves convergence and mitigates overfitting, especially in later epochs (Figure~\ref{fig:eval_log}).
\end{itemize}
These results support prior findings~\cite{swaminathan2015counterfactual,joachims2016counterfactual} and highlight that combining bias-aware weighting with regularization yields more stable offline evaluation and better generalization under unbiased exposure.

\bibliographystyle{ACM-Reference-Format}
\bibliography{main}

\newpage
\appendix

\appendix

\section{Toy Dataset and Experimental Results}
\label{appendix:toy}

\subsection{Toy Dataset Construction}
To illustrate the effect of exposure bias on offline policy evaluation and the comparative behavior of Inverse Propensity Scoring (IPS) and Self-Normalized IPS (SNIPS), we generated a synthetic user–item interaction dataset.

\subsubsection{Ground-Truth Click Probabilities}
We simulated \( U = 1000 \) users and \( I = 200 \) items.  
For each user–item pair \((u, i)\), the ground-truth click-through rate (CTR) was drawn from a Beta distribution:
\[
\mathrm{CTR}_{u,i} \sim \mathrm{Beta}(\alpha=2, \beta=5),
\]
producing a CTR matrix \( \mathbf{C} \in [0,1]^{U \times I} \) that captures heterogeneous user preferences.

\subsubsection{Biased Logging Policy}
To model popularity bias, we defined an exponentially increasing popularity weight:
\[
w_i = \exp\left( \frac{5(i-1)}{I-1} \right), \quad
\pi_{\mathrm{log}}(i) = \frac{w_i}{\sum_{j=1}^{I} w_j},
\]
where \(\pi_{\mathrm{log}}(i)\) is the probability of showing item \(i\) under the logging policy. This heavily favors high-index (popular) items.

\subsubsection{Observed Interaction Generation}
For each user \(u\), we simulated \(K = 5\) exposures by sampling items from \(\pi_{\mathrm{log}}\).  
Given an item \(i\), the click outcome \(y_{u,i}\) was drawn as:
\[
y_{u,i} \sim \mathrm{Bernoulli}(\mathrm{CTR}_{u,i}).
\]
The logging propensity \(\pi_{\mathrm{log}}(i)\) was recorded for each interaction to enable IPS and SNIPS evaluation.  
The resulting dataset contained \( U \times K = 5000 \) exposure events.

\subsection{Results on Toy Dataset}

\subsubsection{Empirical Click-Through Rate per Item}
\begin{figure}[htb]
    \centering
    \includegraphics[width=0.85\linewidth]{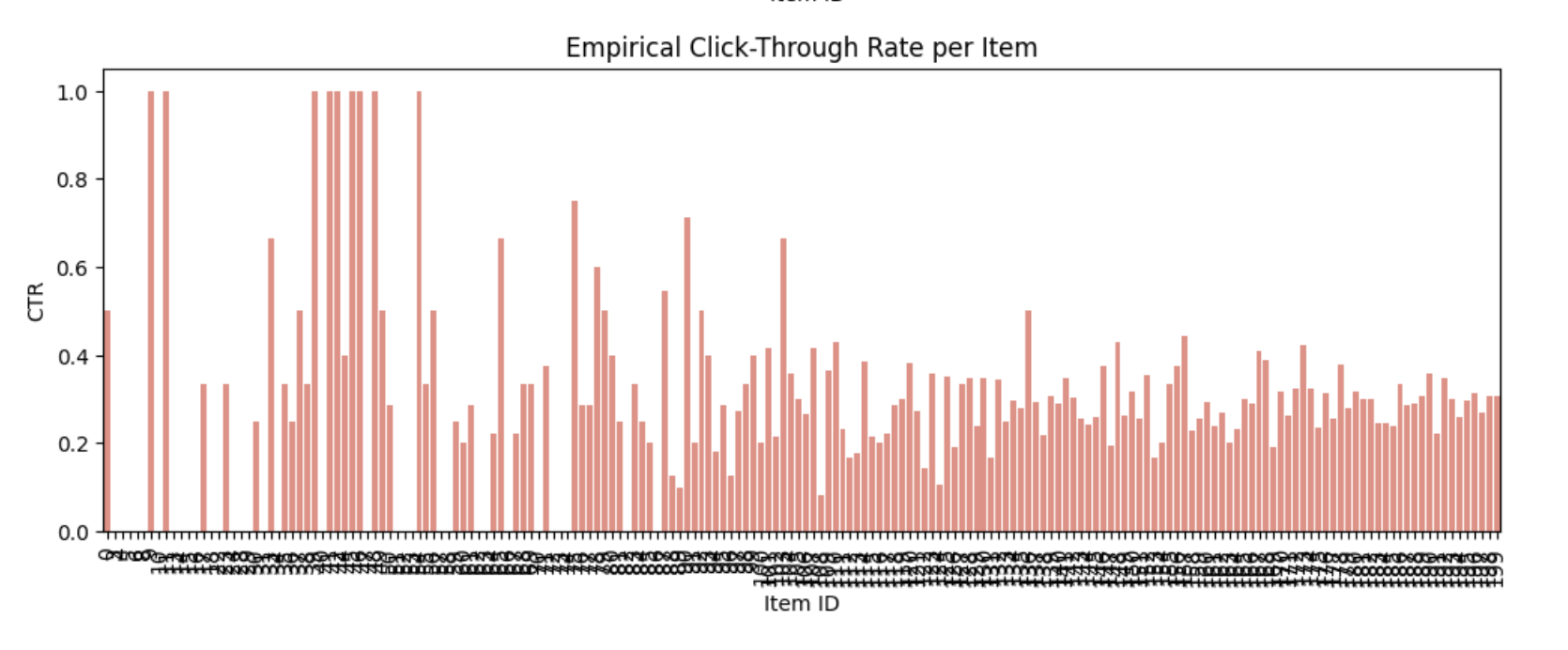}
    \caption{Empirical Click-Through Rate (CTR) per item in the toy dataset. Items with few exposures exhibit inflated CTR variance.}
    \label{fig:click_rate_toy}
\end{figure}

Figure~\ref{fig:click_rate_toy} shows the empirical CTR for each item, computed as the ratio of clicks to exposures.  
Large variance is observed for items with low exposure counts, where CTR estimates are unreliable.

\subsubsection{Distribution of Policy Value Estimates}
\begin{figure}[htb]
    \centering
    \includegraphics[width=0.75\linewidth]{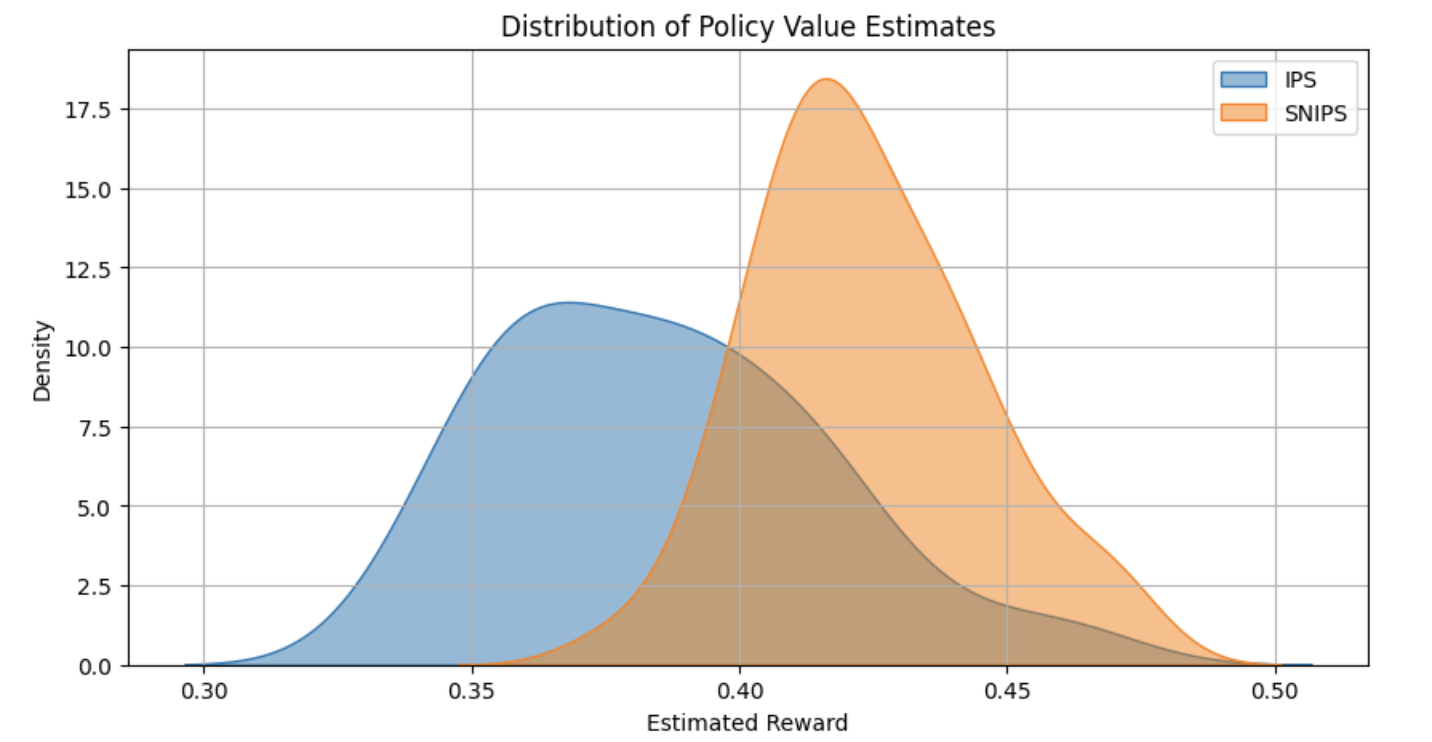}
    \caption{Distribution of policy value estimates for IPS and SNIPS. SNIPS reduces variance compared to IPS.}
    \label{fig:policy_toy}
\end{figure}

Figure~\ref{fig:policy_toy} compares the distribution of policy value estimates for the target policy using IPS and SNIPS.  
SNIPS produces a narrower distribution with reduced variance, validating its robustness to propensity scale variation.

\subsubsection{Item Exposure Count under Biased Logging Policy}
\begin{figure}[htb]
    \centering
    \includegraphics[width=0.85\linewidth]{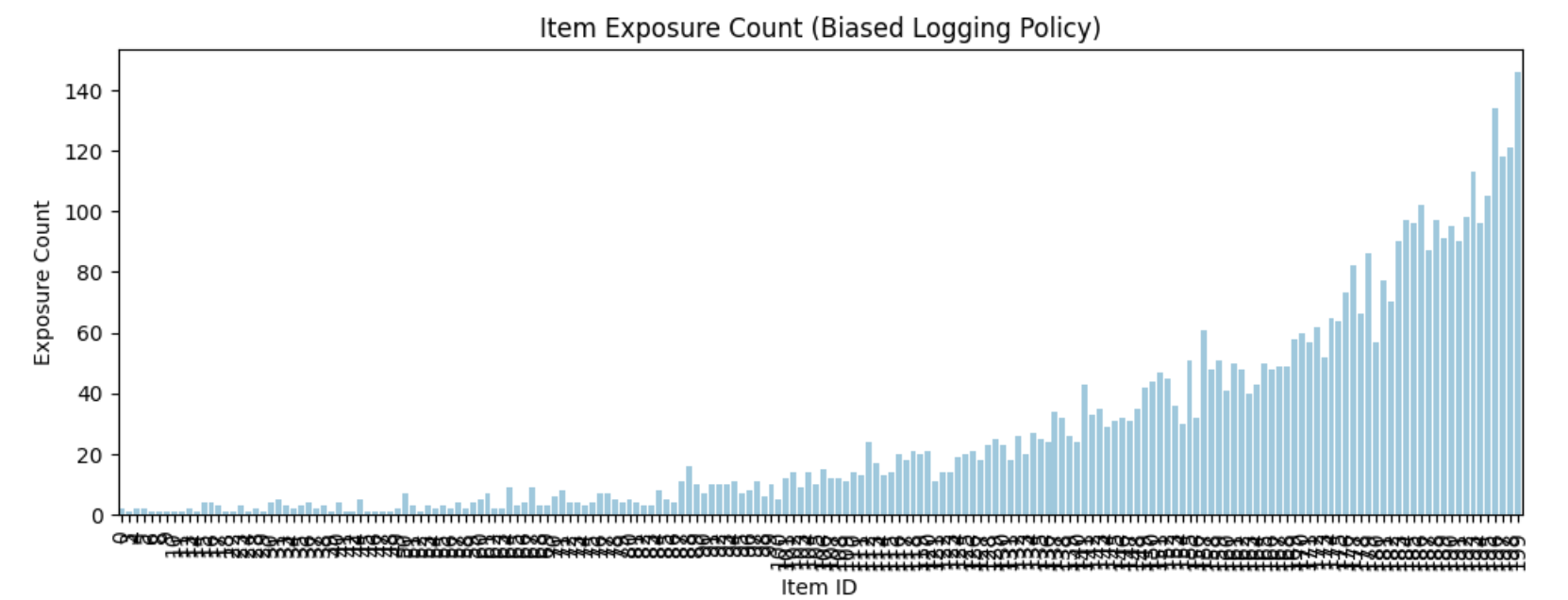}
    \caption{Item exposure counts under the biased logging policy. Popular items dominate exposure frequency.}
    \label{fig:exposure_toy}
\end{figure}

Figure~\ref{fig:exposure_toy} shows exposure counts per item. The skew confirms that the logging policy strongly favors popular items, leading to severe exposure imbalance.

\subsection{Key Observations}
From the toy dataset experiment, we note:
\begin{enumerate}
    \item \textbf{Exposure Bias:} The logging policy induces a heavy-tailed exposure distribution.
    \item \textbf{CTR Variability:} Low-frequency items have unstable CTR estimates.
    \item \textbf{SNIPS Stability:} SNIPS reduces estimation variance, improving reliability in biased exposure settings.
\end{enumerate}

\section{MovieLens Dataset and Descriptive Statistics}
\label{appendix:movielens}

\subsection{Dataset Description}
In addition to the synthetic toy dataset, we conducted experiments on the \textit{MovieLens} dataset to validate our observations in a real-world setting.  
We used the \texttt{MovieLens-1M} subset, which contains approximately 1 million ratings from 6{,}000 users on 4{,}000 movies. To adapt the dataset for click-based evaluation:
\begin{itemize}
    \item Ratings were binarized into \emph{click} events by thresholding at 4 stars (\(\geq 4\) considered a click).
    \item The resulting interaction log contains user–item pairs along with binary click indicators.
    \item No explicit propensity scores are available; instead, logged frequencies serve to illustrate exposure imbalance.
\end{itemize}

\subsection{Empirical Analysis of Logged Data}

\subsubsection{Item Click-Through Rates}
\begin{figure}[htb]
    \centering
    \includegraphics[width=0.95\linewidth]{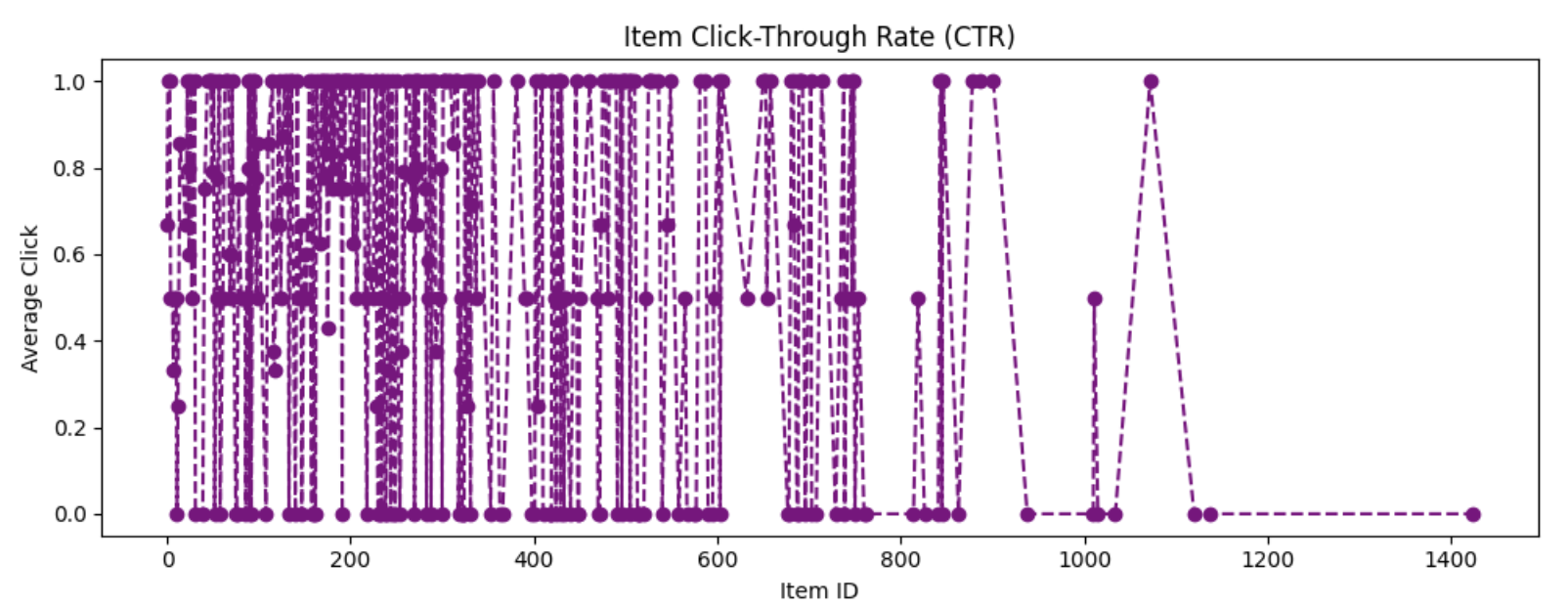}
    \caption{Average item click-through rate (CTR) in the MovieLens dataset. Many items have either very high or very low engagement levels.}
    \label{fig:ctr_movie}
\end{figure}

Figure~\ref{fig:ctr_movie} shows the average click-through rate (CTR) per item. The CTR is computed as the ratio of the number of clicks to the number of exposures for each item.  
The distribution reveals that some items have CTRs close to 1.0 (highly engaging), while a significant number have CTRs near zero.

\subsubsection{Item Exposure Frequency}
\begin{figure}[htb]
    \centering
    \includegraphics[width=0.95\linewidth]{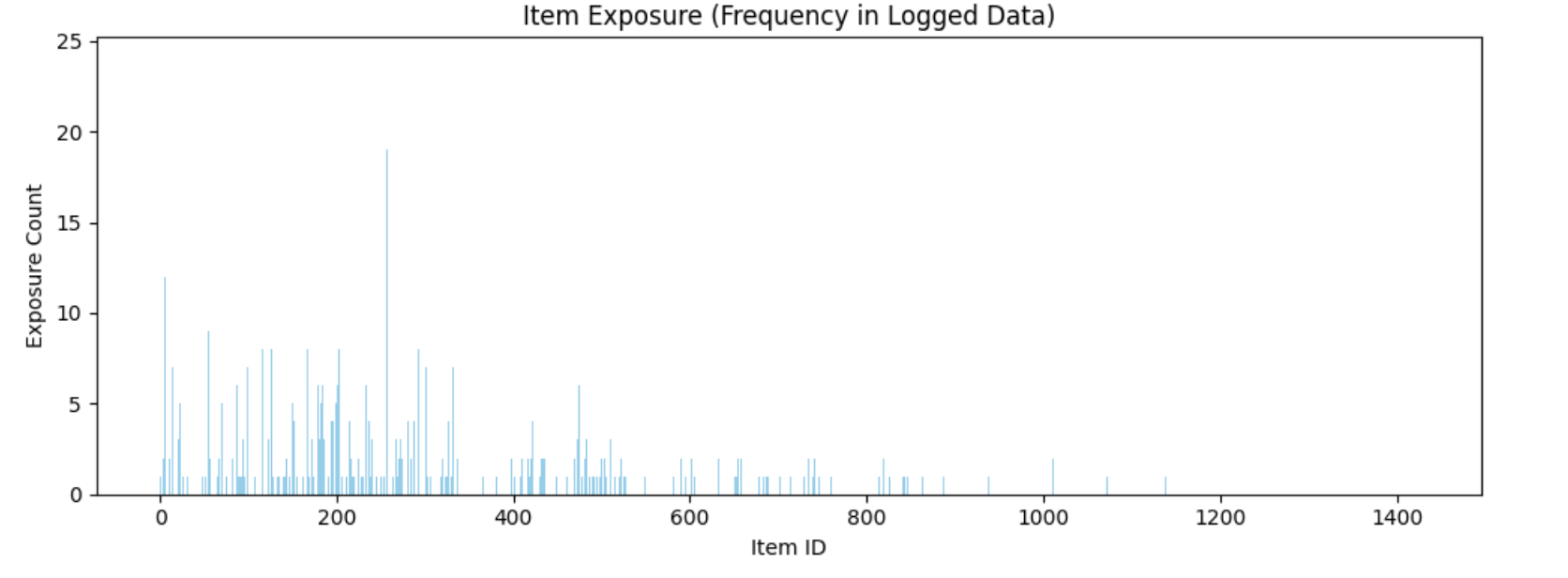}
    \caption{Item exposure counts in the MovieLens dataset. The distribution is highly skewed toward a small subset of popular items.}
    \label{fig:item_movie}
\end{figure}

Figure~\ref{fig:item_movie} shows the frequency of item exposures in the logged data. A long-tail pattern is evident, with a small number of items receiving the majority of exposures, while many items have very few.

\subsubsection{User Interaction Counts}
\begin{figure}[htb]
    \centering
    \includegraphics[width=0.95\linewidth]{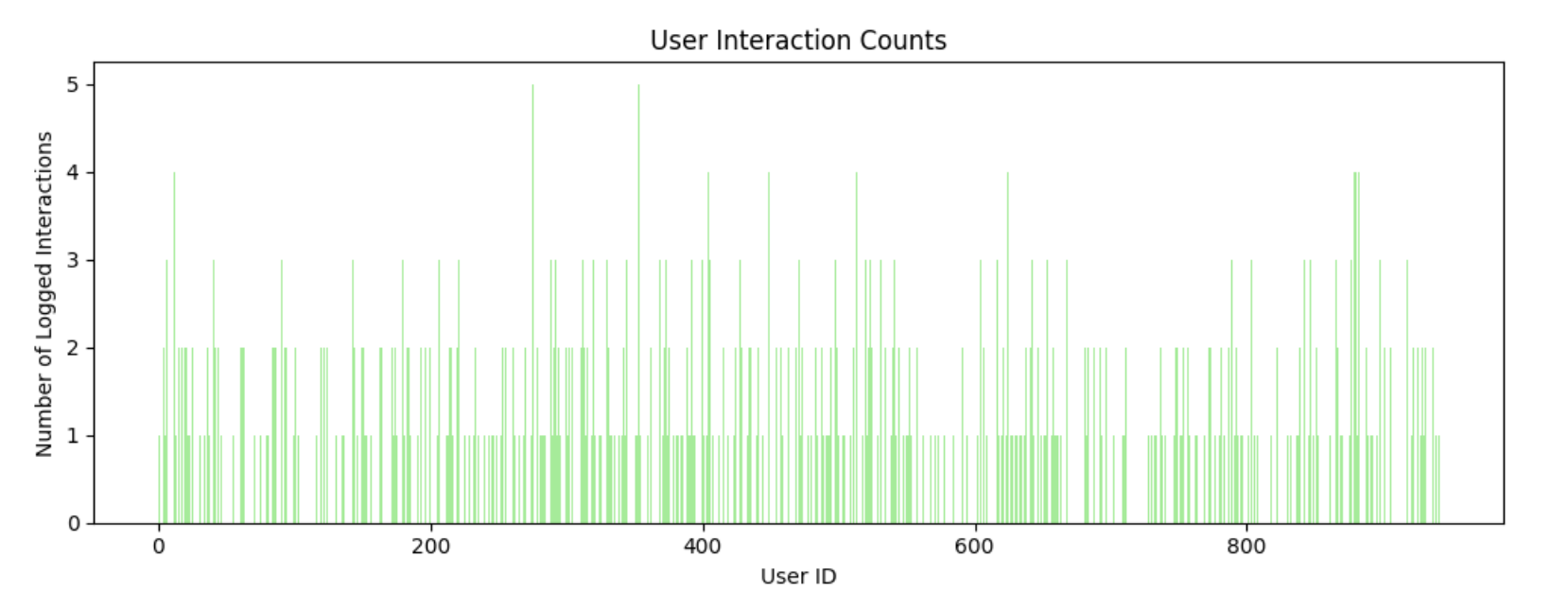}
    \caption{Number of logged interactions per user in the MovieLens dataset. Activity levels vary, with some highly active users.}
    \label{fig:user_movie}
\end{figure}

Figure~\ref{fig:user_movie} displays the number of logged interactions per user. While most users have a moderate number of interactions, the distribution still shows variability, with some users being far more active.

\subsection{Observations}
From the MovieLens dataset statistics, we observe:
\begin{enumerate}
    \item \textbf{Item Popularity Bias:} Exposure frequencies are highly imbalanced across items.
    \item \textbf{Click Rate Variance:} Items with low exposure counts show volatile CTR estimates.
    \item \textbf{User Activity Skew:} A subset of users contributes disproportionately to the total interaction volume.
\end{enumerate}
These patterns resemble those found in the synthetic toy dataset, but with additional complexity from real-world user behavior.

\end{document}